\documentclass{article}
\usepackage{ijcai26}

\usepackage{times}
\usepackage{soul}
\usepackage{url}
\usepackage[hidelinks]{hyperref}
\usepackage[utf8]{inputenc}
\usepackage[small]{caption}
\usepackage{graphicx}
\usepackage{subcaption}
\usepackage{amsmath}
\usepackage{amsthm}
\usepackage{booktabs}
\usepackage{algorithm}
\usepackage{algorithmic}
\usepackage[switch]{lineno}
\usepackage{multirow}
\usepackage{amsfonts}
\usepackage{bbm}
\usepackage{rotating}

\title{Rethinking LIME for Document Classification: The Role of Segmentation in Explanation Stability}
\title{Segmentation is the Hidden Variable of LIME for Document Images}
\title{When Superpixels Fail on Documents: \\A Study of Segmentation for LIME Explanations}

\author{
Quentin Telnoff$^{1,2}$
\and
Emanuela Boros$^1$\and
Mickaël Coustaty$^{1}$\And
Robin Jarry$^2$\And
Fabrice Crohas$^2$\And
Antoine Doucet$^1$\\
\affiliations
$^1$L3i, University of La Rochelle \\
$^2$Itesoft
\emails
\{quentin.telnoff, emanuela.boros, mickael.coustaty, antoine.doucet\}@univ-lr.fr,
\{robin.jarry, fabrice.crohas\}@itesoft.com
}

\begin{document}

\maketitle


\begin{abstract}
Post-hoc explanation methods are widely used to inspect image classifiers, but their reliability depends on design choices that are often treated as implementation details. We study this issue for LIME on document image classification, focusing on the segmentation step that defines the interpretable units being perturbed. Standard image-based LIME typically relies on natural-image superpixels, which are poorly aligned with document structure such as text regions, layout blocks, and identification codes. Using RVL-CDIP, we compare Quickshift and SLIC with document-aware segmentations based on OCR bounding boxes and regular grids. Our results show that segmentation strongly affects explanation consistency, correctness, and local fidelity. Document-aware segmentations produce more stable and faithful explanations, require fewer perturbations to converge, and expose shortcut behaviour based on document identification codes, a known RVL-CDIP bias that superpixel-based LIME often obscures. These findings show that reliable post-hoc explanation requires domain-aware interpretable representations, and that segmentation should be treated as part of the explanation method rather than as neutral preprocessing.
\end{abstract}

\section{Introduction}

\begin{figure}[htbp]
     \centering
     \begin{subfigure}[b]{0.43\textwidth}
         \centering
         \includegraphics[width=\textwidth]{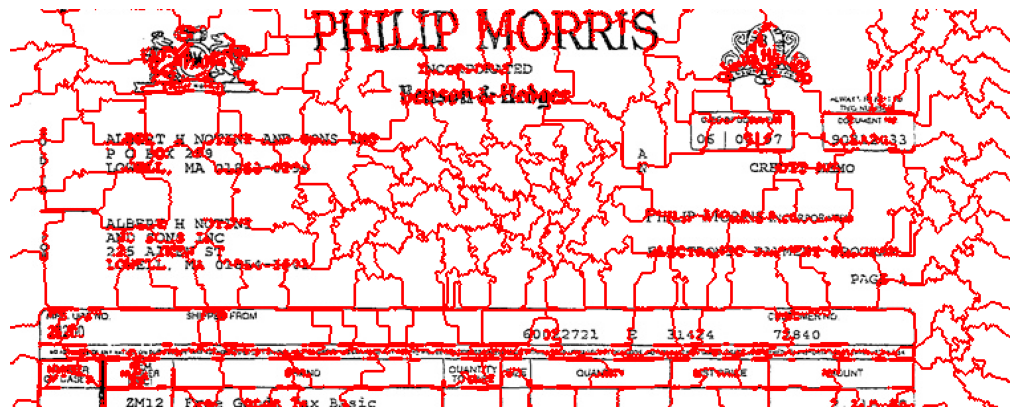}
         \caption{Quickshift \cite{vedaldi2008quick}}
         \label{fig:quickshift_doc}
     \end{subfigure}
     \hfill 
     \begin{subfigure}[b]{0.43\textwidth}
         \centering
         \includegraphics[width=\textwidth]{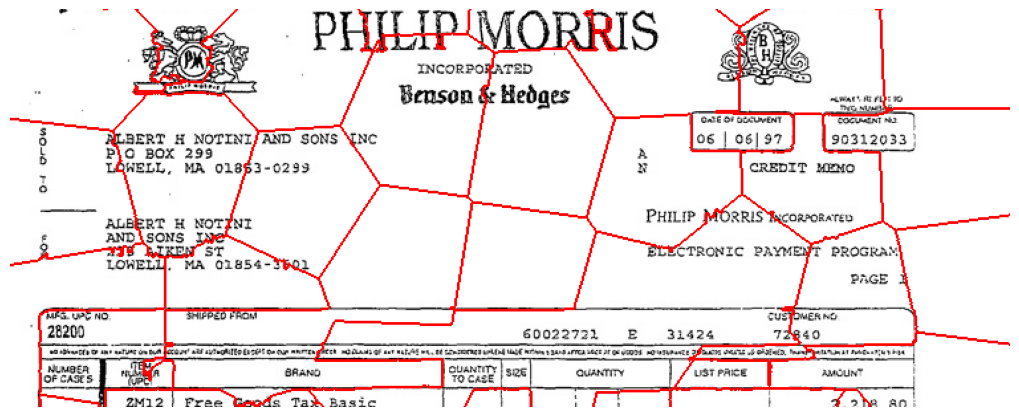}
         \caption{SLIC \cite{achanta2012slic}}
         \label{fig:slic_doc}
     \end{subfigure}
     
     \caption{Standard superpixel segmentation fails to capture meaningful structural components in document images.}
     \label{fig:quickshift_slic_on_document}
\end{figure}

Explainable artificial intelligence (XAI) has become an important requirement for deploying machine learning systems in high-stakes settings. In document AI, classification models are increasingly used for archival categorization, administrative workflow automation, and large-scale record processing. Although deep neural networks achieve strong predictive performance, their decisions remain difficult to interpret, limiting trust and error analysis in practice~\cite{salih2025perspective}.

Among post-hoc explanation methods, LIME remains widely used due to its model-agnostic formulation and intuitive local surrogate mechanism~\cite{zhao_explainability_2024}. In image classification, LIME partitions the image into regions via segmentation, then perturbs their presence or absence to estimate local feature importance~\cite{ribeiro_why_2016}. However, standard implementations rely on algorithms such as SLIC~\cite{achanta2012slic} or Quickshift~\cite{vedaldi2008quick}, designed for natural images, which group pixels by low-level visual similarity rather than structural document properties such as text blocks, columns, or tables. As shown in Figure~\ref{fig:quickshift_slic_on_document}, this results in segmentations that fragment the page layout instead of capturing coherent document regions on a document from RVL-CDIP~\cite{harley2015icdar}.

This matters because segmentation directly defines the dimensionality, geometry, and semantic coherence of the features on which LIME builds its explanations. Yet its role has received comparatively little attention for structured visual data~\cite{knab_beyond_2025}. We therefore ask: \emph{how much does the choice of segmentation affect the stability and reliability of LIME explanations on document images?} 

We compare superpixel-based methods with document-aware alternatives derived from OCR regions and regular grids on RVL-CDIP, and make three contributions:
\begin{enumerate}
    \item We show that segmentation is a central design choice in image-based LIME, because it defines the interpretable representation on which post-hoc explanations are computed.
    \item We evaluate the reliability of LIME explanations across segmentation strategies using consistency, correctness, and local fidelity metrics.
    \item We show that document-aware segmentation improves not only explanation stability, but also bias discovery, by revealing RVL-CDIP shortcut behaviour based on document identification codes that superpixel-based LIME often obscures.
\end{enumerate}

\section{Related Work}

\begin{figure*}[ht]
   \centering
   \includegraphics[width=.9\textwidth]{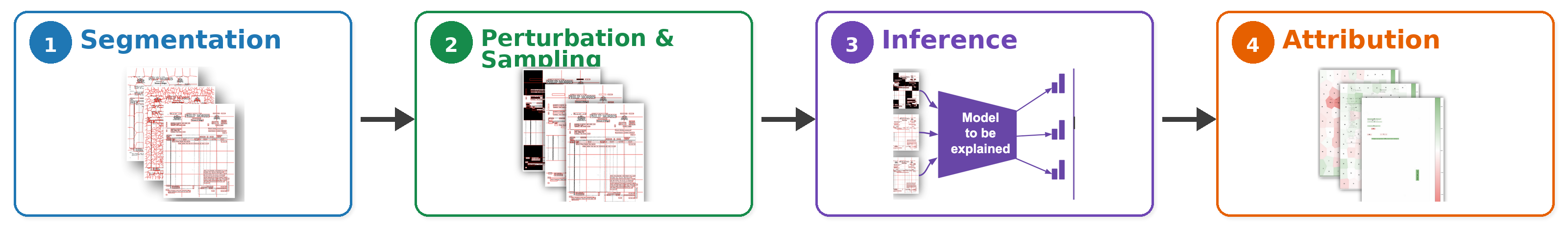}
\caption{Image-based LIME pipeline. Segmentation defines the interpretable regions, perturbations generate neighborhood samples, and a local surrogate model produces region-level attributions.}
   \label{fig:lime_pipeline_overview}
\end{figure*}

Explainability for document image classification lies at the intersection of several research areas: document AI, post-hoc explanation methods, image-based perturbation techniques, and the evaluation of explanation quality.

\paragraph{Explainability for Document AI.}
Explainability in document AI is receiving growing attention, but the field remains less mature than explainability for natural images or tabular data~\cite{saifullah_docxplain_2024,saifullah_reality_2024}. Existing studies show that standard explainability methods often perform poorly on document image classification and may fail to capture the document-specific evidence used by models~\cite{saifullah_reality_2024}. This is particularly important because document images are not natural scenes: their meaning is often encoded through layout, textual regions, background structure, tables, margins, and other spatial regularities. Many document explanation pipelines still adapt generic computer vision tools with limited consideration for these document-specific properties~\cite{saifullah_docxplain_2024,larson_evaluation_2023,larson_spurious_2025}. This creates a gap between how explanations are generated and how documents are actually structured and interpreted.

\paragraph{LIME and Its Variants.}
LIME is one of the most widely used local post-hoc explanation methods. Its central idea is to approximate the behaviour of a black-box model in the neighbourhood of a single instance by fitting a simple interpretable surrogate model on perturbed samples. Because of its model-agnostic formulation, LIME has been applied across a wide range of modalities, including text, tabular data, and images~\cite{band2023application,mardaoui2021analysis,garreau2020looking}. A substantial body of work has since examined limitations of LIME and proposed variants that modify individual components of the method, including the sampling strategy, neighbourhood definition, surrogate formulation, and weighting mechanism~\cite{zhou_s-lime_2021,zafar_deterministic_2021,zhao_baylime_2021,tan_glime_2023}. These studies show that explanation quality can be sensitive to implementation choices, but they typically focus on improving one component at a time.

\paragraph{Segmentation in Image-based LIME.}
In image-based uses of LIME, segmentation is the step that defines the interpretable units on which perturbations are performed. Standard choices such as SLIC~\cite{achanta2012slic} and Quickshift~\cite{vedaldi2008quick} were originally developed for natural images, where grouping pixels by local visual similarity can be a reasonable approximation of semantic structure. However, this assumption does not necessarily hold for document images, whose organisation is driven less by texture and object boundaries than by layout regularities such as text blocks, columns, tables, margins, and background regions. Recent work has started to question the role of segmentation in explanation quality and to explore alternatives to standard superpixel pipelines~\cite{schallner_effect_2019,knab_beyond_2025,pihlgren_segmentation_2025}. These studies suggest that segmentation can substantially affect explanation quality, but they mostly focus on natural-image settings or generic image explanation pipelines. For document images, where the relevant structure is often layout-based rather than object-based, segmentation remains a central but under-discussed design choice~\cite{saifullah_docxplain_2024,saifullah_reality_2024}.

\paragraph{Stability and Evaluation of Explanations.}
A recurring criticism of local explanation methods is their instability: repeated runs with the same model and input may produce different explanations, especially when perturbation-based sampling is involved~\cite{amparore_trust_2021,visani_statistical_2020}. This issue has motivated more robust variants of LIME and broader discussions about what should count as a reliable explanation~\cite{zhou_s-lime_2021,zafar_deterministic_2021,zhao_baylime_2021,tan_glime_2023}. At the same time, the evaluation of XAI methods remains highly fragmented. Different studies rely on different notions of quality, such as fidelity, stability, faithfulness, sparsity, correctness, or human usefulness, often without a common protocol~\cite{nauta_anecdotal_2023,knab_which_2025,alangari_exploring_2023}. As a result, comparing findings across papers is difficult, and it remains unclear which design choices matter most in practice.




Our work connects these strands by studying LIME for document image classification through the specific lens of segmentation. Rather than proposing a new explanation method, we ask whether the segmentation used to define LIME's interpretable space can explain a substantial part of its instability on document images. In doing so, we address the gap between generic image explanation pipelines and the layout-driven structure of document images, and we evaluate this question through consistency and correctness metrics.


\section{Method}\label{sec:methodology}

This section formalizes the role of segmentation in image-based LIME and describes the segmentation strategies evaluated in this work as shown in Figure~\ref{fig:lime_pipeline_overview}. 

\subsection{Problem Setting}\label{sec:problem-setting}
Let $f: \mathcal{X} \rightarrow \mathcal{Y}$ be a document image classifier, where $\mathcal{X}$ denotes the space of document images and $\mathcal{Y}$ the class probabilities. Given an input image $x \in \mathcal{X}$ and a predicted class of interest, LIME explains the prediction by constructing a local interpretable model around $x$ based on perturbed versions of the input~\cite{ribeiro_why_2016}. In the image setting, the LIME methodology is divided into three steps.

\paragraph{Segmentation.} The original image is partitioned into patches called super-pixels. Let $\mathcal{S}$ be a segmentation algorithm. For each pixel $p \in x$, there exists a unique index $i \in \{1, \ldots, d\}$ such that $\mathcal{S}(p) = i$, where $d$ is the total number of super-pixels, which depends on both the input image and the chosen algorithm.

\paragraph{Perturbation and sampling.} To capture the behaviour of $f$ in the neighbourhood of $x$, data are generated around $x$ by selectively keeping certain super-pixels visible while masking others with a predefined reference colour $c$. Formally, a binary vector $z \in \{0, 1\}^d$ is sampled according to a specified probability distribution $\mathcal{L}$. A perturbed image $\tilde{x} \in \mathcal{X}$ is then constructed by retaining the super-pixels where $z_i = 1$ and occluding those where $z_i = 0$ with the reference colour $c$.

\paragraph{Attribution.} Under a given segmentation and perturbation strategy, LIME can be framed as a local weighted regression problem in the interpretable space, i.e., the regression is performed on $\{z^{(i)}\}_i$ rather than directly on the image.

A family of surrogate models $G$ is first chosen (e.g., linear models or decision trees). A dataset of $N$ perturbed samples, denoted $\mathcal{D}$ and defined in Equation~\ref{eq:perturbationdata}, is then generated, mapped back to perturbed images, and passed through the classifier $f$:
\begin{equation}\label{eq:perturbationdata}
    \mathcal{D} = \left\{ \left(z^{(i)},\, p^{(i)}\right)_{i=1}^{N} \;\middle|\; z^{(i)} \sim \mathcal{L},\; f\!\left(\tilde{x}^{(i)}\right) = p^{(i)} \in (0, 1) \right\}.
\end{equation}
A locality weighting function $\pi$ is defined in Equation~\ref{eq:weightedequation} as:
\begin{equation}\label{eq:weightedequation}
\pi \colon \begin{array}{ccl}
\{0,1\}^d & \longrightarrow & \mathbb{R} \\
z & \longmapsto & \exp\!\left(\dfrac{-D(\mathbf{1}_d,\, z)}{\sigma^2}\right),
\end{array}
\end{equation}
where $D(\mathbf{1}_d, z)$ denotes the distance between the all-ones vector $\mathbf{1}_d$ and $z$, and $\sigma^2 > 0$ is a bandwidth hyperparameter controlling the locality of the kernel.

\begin{figure*}[htbp] 
    \centering
    \begin{subfigure}{0.22\textwidth}
        \includegraphics[width=\linewidth]{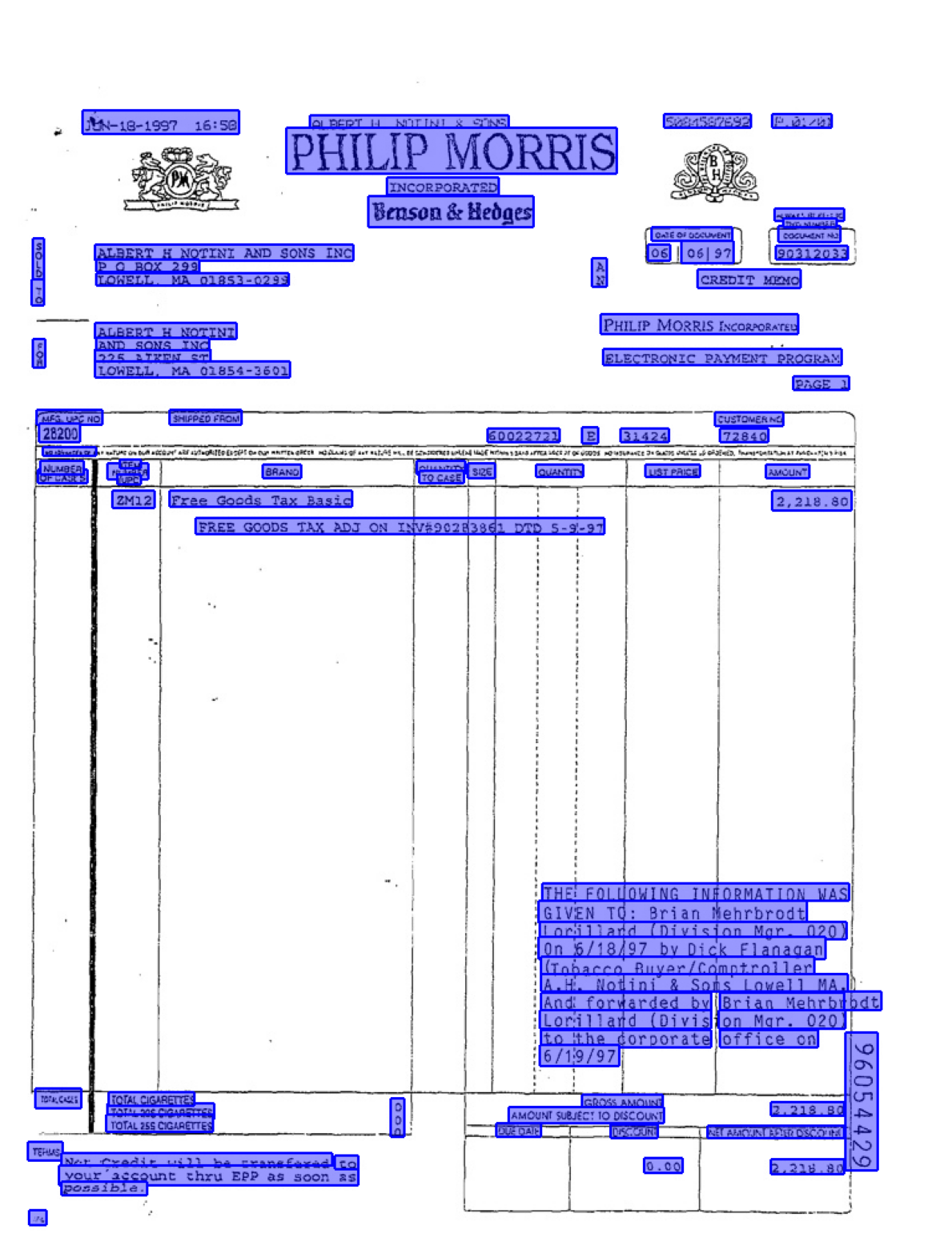}
        \caption{ }
        \label{fig:ocronly_seg}
    \end{subfigure}\hfill
    \begin{subfigure}{0.22\textwidth}
        \includegraphics[width=\linewidth]{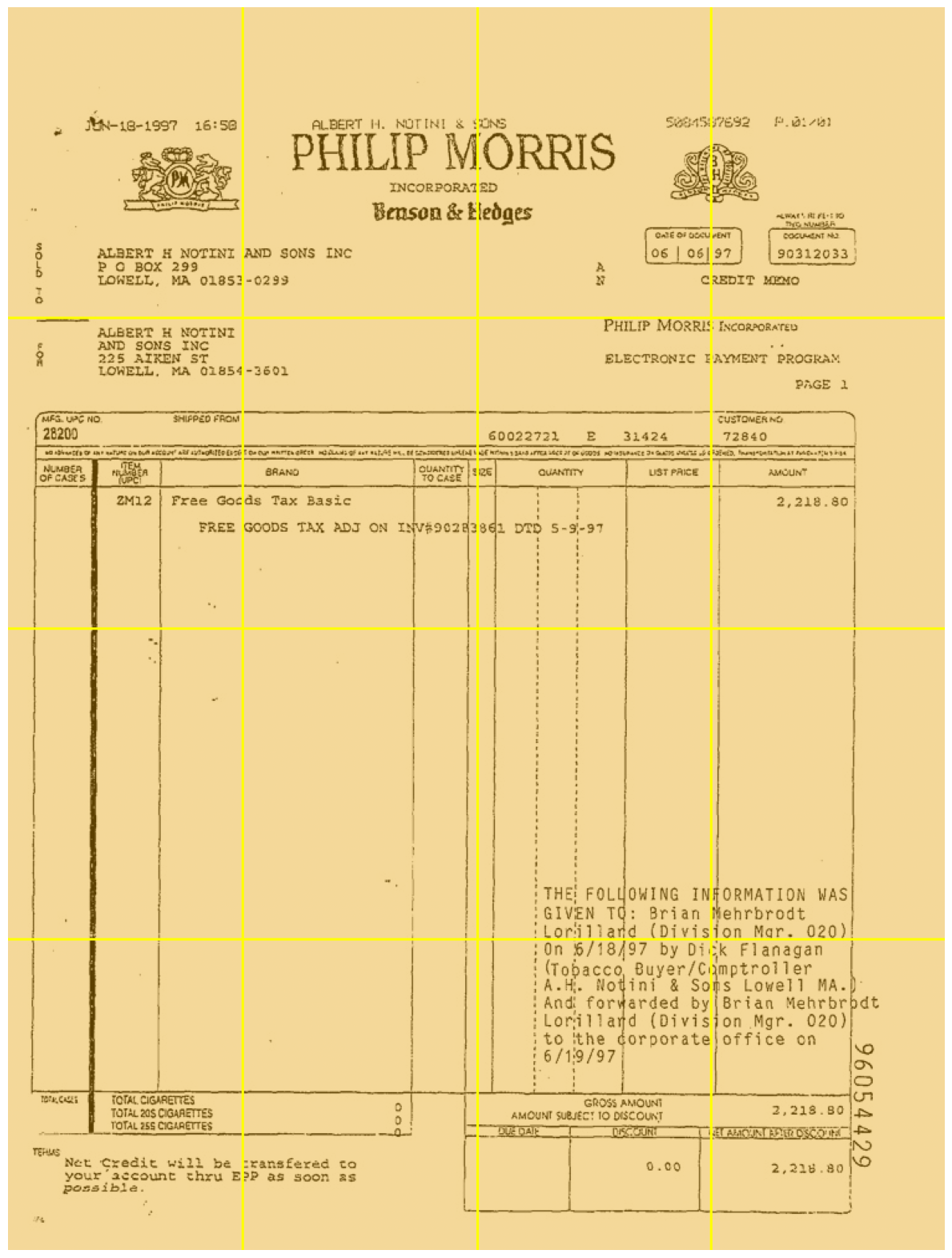}
        \caption{ }
        \label{fig:gridonly_seg}
    \end{subfigure}\hfill
    \begin{subfigure}{0.22\textwidth}
        \includegraphics[width=\linewidth]{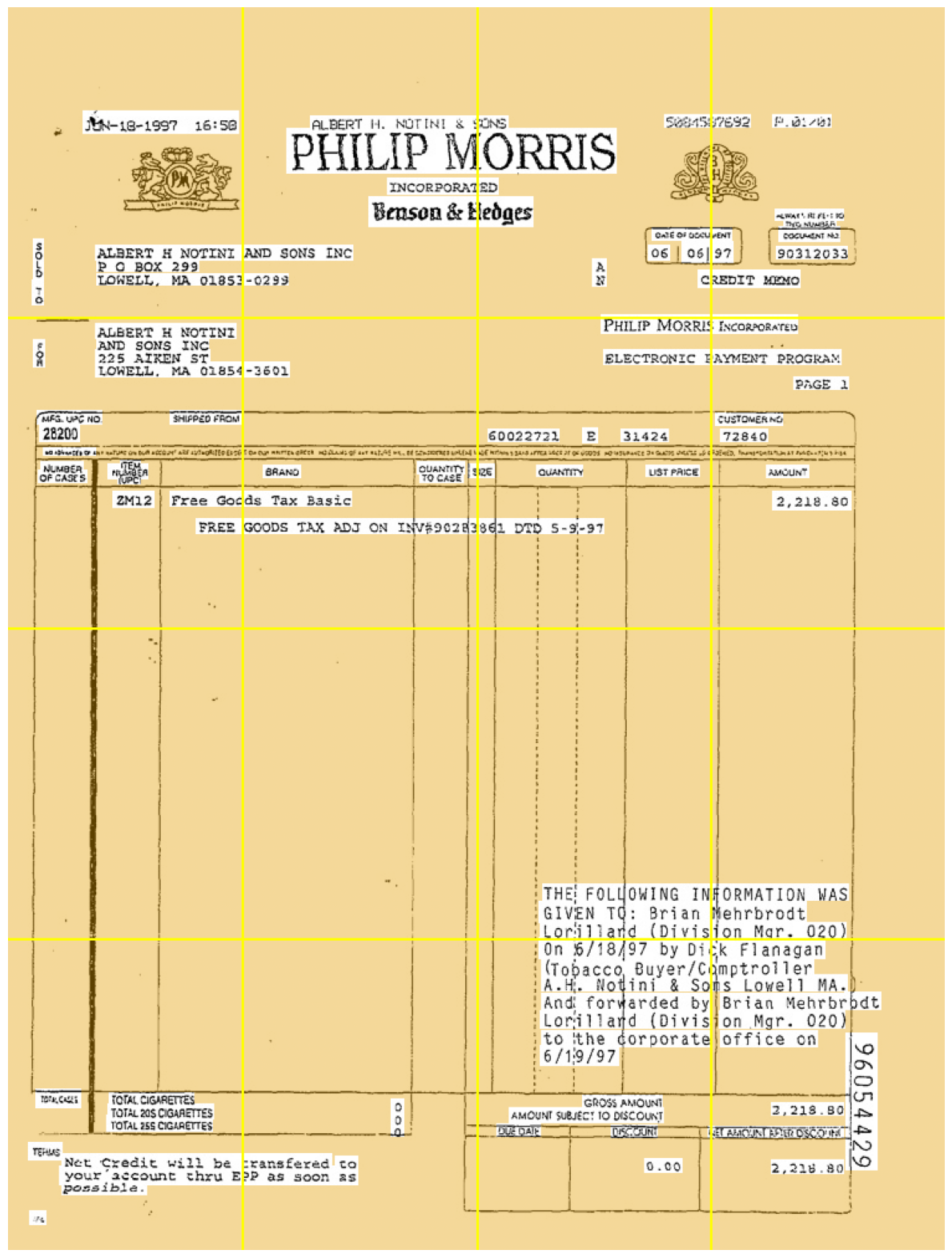}
        \caption{ }
        \label{fig:grid-ocr_seg}
    \end{subfigure}\hfill
    \begin{subfigure}{0.22\textwidth}
        \includegraphics[width=\linewidth]{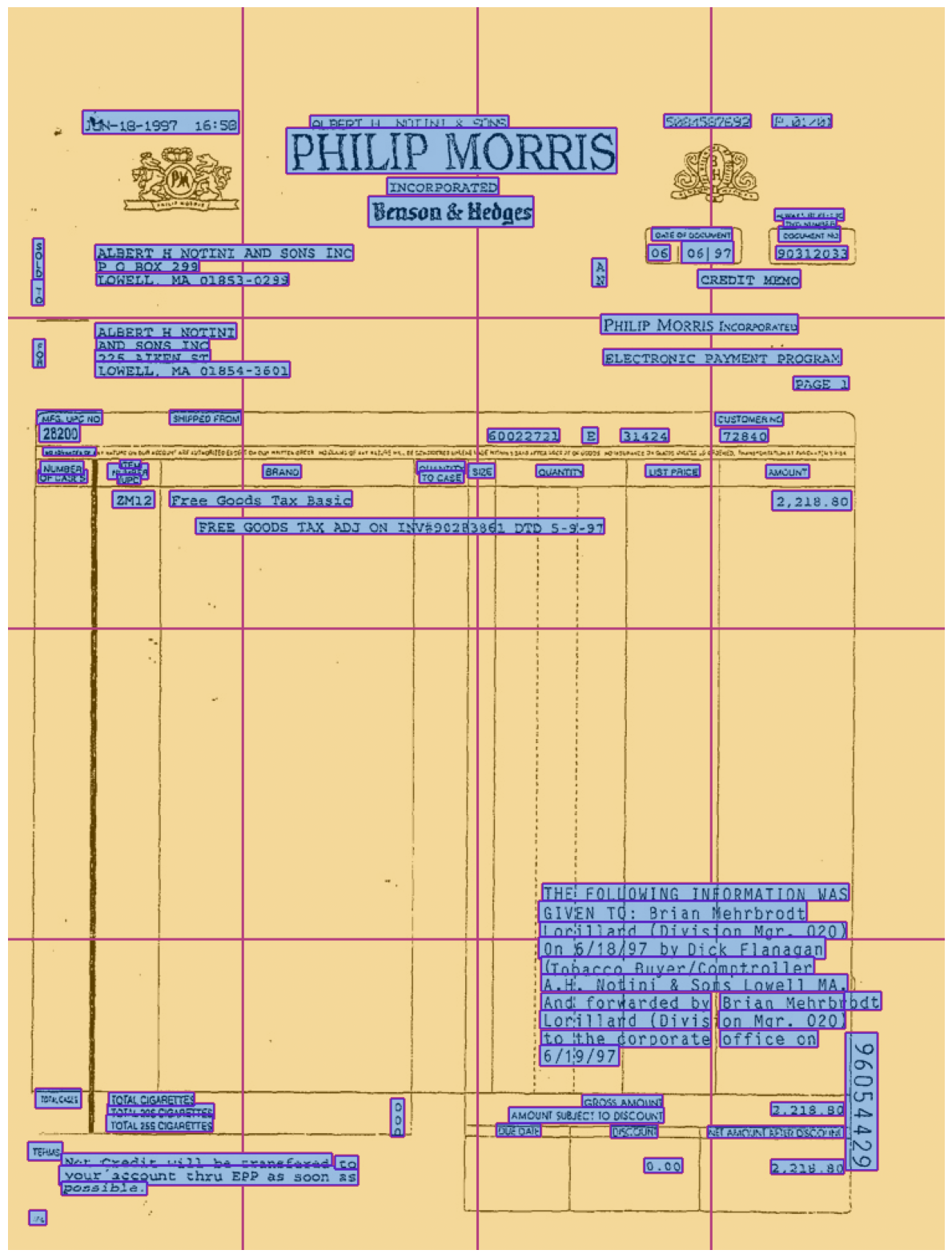}
        \caption{ }
        \label{fig:ocr&grid}
    \end{subfigure}
    
    \caption{Various stages in the construction of the document-aware segmentation. Only the highlighted content is used in the segmentation.}
    \label{fig:document-aware segmentation}
\end{figure*}

A surrogate model $g \in G$ is then fitted on $\mathcal{D}$ by solving the optimization problem in Equation~\ref{eq:optimizationproblem}:
\begin{equation}\label{eq:optimizationproblem}
\xi(x) = \arg\min_{g \in G} \sum_{(z,\, p)\,\in\,\mathcal{D}} \pi(z) \Big(g(z) - p\Big)^2 + \Omega(g),
\end{equation}
where $\pi$ is the locality kernel defined above and $\Omega(g)$ controls the complexity of the surrogate model.





\subsection{Why Segmentation Matters}

Segmentation is not a neutral preprocessing step. In the notation of Section~\ref{sec:problem-setting}, the segmentation function \(\mathcal{S}\) defines the \(d\) interpretable regions of the image, and therefore determines the coordinates of the perturbation vector \(z \in \{0,1\}^d\). Consequently, changing \(\mathcal{S}\) changes the explanation problem itself.

First, \(\mathcal{S}\) defines the \textbf{semantic granularity} of the interpretable units: a region may correspond to a coherent document component, such as a text block, or to an arbitrary visual fragment. Second, it determines the \textbf{dimensionality} \(d\) of the surrogate regression problem. Third, it affects the \textbf{plausibility of perturbations}, since setting \(z_i=0\) masks one region, but the meaning of that masking operation depends on whether the region is structurally meaningful.

This is especially important for document images. Unlike natural scenes, documents are highly structured visual objects whose meaning is often encoded by layout, alignment, whitespace, and textual organization. A segmentation \(\mathcal{S}\) that ignores this structure may produce perturbations \(\tilde{x}\) that are formally valid for LIME but poorly matched to the document domain.

\subsection{Segmentation Strategies}\label{sec:segmentation-strategies}

We compare two broad families of segmentation strategies. Each strategy corresponds to a different segmentation function \(\mathcal{S}\), and therefore induces a different set of interpretable regions and a different dimensionality \(d\) for the perturbation space \(\{0,1\}^d\).

\paragraph{Superpixel-based segmentations.}
We first consider standard image segmentation methods commonly used with LIME. In this setting, \(\mathcal{S}\) partitions the image into visually homogeneous superpixels, which are then used as the binary features perturbed by LIME.
\begin{enumerate}
    \item \textbf{Quickshift} approximates kernelized mean-shift within a 5D spatial-colour feature space to compute multiscale hierarchical segmentations~\cite{vedaldi2008quick}.
    \item \textbf{SLIC} performs K-means in the 5D space of colour information and image location, producing compact superpixels that are commonly used as interpretable units in image-based LIME~\cite{achanta2012slic}.
\end{enumerate}

\paragraph{Document-aware segmentations.}
We then investigate segmentation functions \(\mathcal{S}\) designed to reflect the structure of document images. Instead of defining regions only from low-level visual similarity, these strategies define the coordinates of \(z \in \{0,1\}^d\) using textual zones, spatial layout, and coarse page structure. The complete segmentation pipeline is illustrated in Figure~\ref{fig:document-aware segmentation}.

First, we extract textual regions using an optical character recognition (OCR) system~\cite{cui2025paddleocr}. This produces a segmentation in which the interpretable regions correspond to OCR bounding boxes, denoted as \textbf{BBs only} and visualized in Figure~\ref{fig:ocronly_seg}. To capture the global spatial layout, we also introduce a uniform \(n_r \times n_c\) grid over the document, denoted as \textbf{Grid \(n_r \times n_c\) w/o BBs} and shown in Figure~\ref{fig:gridonly_seg} for \(n_r=4\) and \(n_c=4\).

Next, to combine textual and layout information without fragmenting OCR regions, we subtract the OCR bounding boxes from the regular grid. This intermediate representation is shown in Figure~\ref{fig:grid-ocr_seg}. Finally, the full document-aware segmentation, denoted as \textbf{BBs w/ Grid}, is obtained by taking the union of the intact OCR bounding boxes and the remaining grid regions, as shown in Figure~\ref{fig:ocr&grid}.

These strategies define interpretable regions that are non-overlapping and more closely aligned with document structure. As a result, they change both the semantic meaning of the perturbed coordinates \(z_i\) and the dimensionality \(d\) of the local surrogate problem.

\section{Experimental Setup}\label{sec:experimentalsetup}
This section describes the experimental setup used to study the effect of segmentation on LIME explanations. 

\subsection{Data}
We use RVL-CDIP, a standard benchmark for document image classification. The full dataset comprises 400,000 grayscale images of scanned documents distributed evenly across 16 categories, including letters, forms, emails, invoices, and budgets. Following the standard split, the dataset is partitioned into 320,000 training images, 40,000 validation images, and 40,000 test images~\cite{harley2015icdar}. We use the full training split to train the document classifier that is later explained.

For the explanation experiments, we evaluate on a focused subset of the test set rather than on all 16 classes. Specifically, we select three RVL-CDIP classes: \textit{Invoice}, \textit{Budget}, and \textit{Form}. We choose these classes because they are visually and structurally close: they often share similar layouts, including tabular regions, aligned text blocks, and form-like organization. Moreover, the model's confusion matrix shows frequent confusions among these classes, indicating that their visual similarity also affects classification behaviour. This makes them particularly relevant for studying the role of segmentation in document explanations. 


\subsection{Backbone Model}
We used a ResNet-50 architecture initialized with standard ImageNet pre-trained weights and trained for document classification~\cite{he2016deep}. To adapt the model to our specific task, the final fully connected layer was replaced with a linear layer corresponding to the target number of document classes. We then trained and evaluated it on the full training and testing splits of the RVL-CDIP dataset, respectively\footnote{The model was optimized using the Adam optimizer with an initial learning rate of $1 \times 10^{-4}$ and a batch size of 32. We minimized the standard categorical cross-entropy loss over 10 epochs. Based on these results, we selected the model checkpoint at epoch 3, as the model exhibited overfitting in subsequent epochs.}.

\begin{table*}[ht]
    \centering
    \begin{tabular}{clcccc}
    \toprule
         &&  Consistency & Local Fidelity & \multicolumn{2}{c}{Correctness}\\ 
         \cmidrule(lr){3-3} \cmidrule(lr){4-4} \cmidrule(lr){5-6}
 & Segmentation algorithm & Spearman ($\uparrow$) & $R^2$ ($\uparrow$) & Insertion ($\uparrow$) & Deletion ($\downarrow$) \\
 \midrule
\multirow{2}{*}{\rotatebox{90}{\textbf{SBS}}} 
 & Quickshift & $0.432 \pm 0.062$ & $0.389 \pm 0.040$ & $0.080 \pm 0.049$ & $0.073\pm0.048$ \\
 & SLIC & $0.792 \pm 0.067$ & $0.218 \pm 0.142$ & $0.374 \pm 0.244$ & $0.089\pm0.081$ \\
  \midrule
\multirow{7}{*}{\rotatebox{90}{\begin{tabular}{c}\textbf{Doc.-Aware} \\ \textbf{Segmentation}\end{tabular}}} 
 & BBs only & $0.933 \pm 0.045$ & $0.557 \pm 0.181$ & $0.764 \pm 0.303$ & $0.031\pm0.027$ \\
 & Grid $4\times4$ w/o BBs & $0.975 \pm 0.014$ & $0.508 \pm 0.141$ & $0.589 \pm 0.203$ & $0.126\pm0.118$ \\
 & BBs w/ $1\times1$ Grid & $0.875 \pm 0.083$ & $0.571 \pm 0.226$ & $0.179 \pm 0.137$ & $0.359\pm0.183$ \\
 & BBs w/ $4\times4$ Grid & $0.843 \pm 0.064$ & $0.328 \pm 0.110$ & $0.522 \pm 0.212$ & $0.133\pm0.100$ \\
 & BBs w/ $10\times10$ Grid & $0.763 \pm 0.056$ & $0.244 \pm 0.060$ & $0.389 \pm 0.240$ & $0.084\pm0.058$ \\
 & BBs w/ $20\times10$ Grid & $0.688 \pm 0.051$ & $0.266 \pm 0.052$ & $0.304 \pm 0.208$ & $0.086\pm0.058$ \\
 & BBs w/ $30\times10$ Grid & $0.615 \pm 0.043$ & $0.285 \pm 0.045$ & $0.266 \pm 0.198$ & $0.090\pm0.058$ \\
 \bottomrule
    \end{tabular}
    \caption{Comparison of segmentation strategies for LIME explanations on RVL-CDIP. We report consistency using Spearman correlation, local fidelity using surrogate \(R^2\), and correctness using insertion and deletion AUC. SBS denotes superpixel-based segmentation.}
\label{tab:segmentation_lime_evaluation}
\end{table*}

\subsection{LIME Configurations}
Building upon the notations introduced in Section~\ref{sec:problem-setting}, we detail the hyperparameters chosen for our LIME experiments. 
\begin{itemize}
    \item \textbf{Segmentation step.} The investigated strategies are outlined in Section~\ref{sec:segmentation-strategies}.
    \item \textbf{Perturbation and sampling step.} The reference colour $c$ is set to black, and the probability distribution $\mathcal{L}$ used to sample the binary vectors is the uniform distribution $\mathcal{U}(\{0,1\}^d)$.
    \item \textbf{Attribution step.} The family of surrogate models $G$ is restricted to linear models. For the locality weighting function $\pi$, we use the Euclidean distance for $D$ and set the kernel bandwidth to $\sigma=1$. Regarding the optimization problem, we do not apply any regularization, setting the complexity term $\Omega(g)=0$. Finally, we evaluate the impact of the perturbation dataset size by varying $N := |\mathcal{D}|$ from 100 to 8,000 (using steps of 100 up to 1,000, and steps of 1,000 thereafter).
\end{itemize}


\section{Evaluation Protocol}
\label{sec:evaluation_protocol}

Our goal is to assess whether different segmentation strategies lead to reliable LIME explanations. Because explanation quality cannot be reduced to a single criterion, we evaluate three complementary properties: \textbf{consistency}, \textbf{correctness}, and \textbf{local fidelity}. This allows us to measure whether explanations are stable across runs, whether the highlighted regions affect the classifier output, and whether the local surrogate accurately approximates the classifier.

\subsection{Consistency}

We define consistency as the stability of LIME explanations across repeated runs with the same input image \(x\), classifier \(f\), segmentation algorithm \(\mathcal{S}\), and hyperparameters, but with different perturbation samples~\cite{visani_statistical_2020,amparore_trust_2021}. For a fixed segmentation \(\mathcal{S}\), the interpretable space is \(\{0,1\}^d\), where \(d\) is the number of regions induced by \(\mathcal{S}\). Each independent run \(a\) samples a different perturbation dataset \(\mathcal{D}^{(a)}\) and produces an explanation vector \(\xi^{(a)}(x) \in \mathbb{R}^d\), whose entries correspond to the importance scores of the \(d\) regions.

Since we are interested in whether LIME assigns similar relative importance to the same regions, we measure consistency using Spearman's rank correlation between explanation vectors. For \(R\) independent LIME runs, we report the average pairwise Spearman correlation:
\begin{equation}
    \mathrm{Consistency}(x,\mathcal{S}) =
    \frac{2}{R(R-1)}
    \sum_{1 \leq a < b \leq R}
    \rho_s\!\left(\xi^{(a)}(x),\xi^{(b)}(x)\right).
\end{equation}

Higher values indicate that repeated executions produce similar rankings of the same interpretable regions. Lower values indicate that the explanation is more strongly affected by the random perturbation samples \(\mathcal{D}^{(a)}\).

\subsection{Correctness}
To evaluate the correctness of the explanations, we use \textbf{deletion} and \textbf{insertion} metrics~\cite{petsiuk_rise_2018,nauta_anecdotal_2023}, which test whether the regions identified as important by LIME affect the classifier's score. The process involves ranking the image regions from most to least important and modifying them sequentially.
\begin{enumerate}
    \item \textbf{Deletion:} Starting with the original image, we progressively mask the regions from most to least important and measure the model's output at each step. A faithful explanation should cause the model's confidence to drop rapidly.
    \item \textbf{Insertion: }Starting with a fully masked baseline image, we progressively restore the regions from most to least important. In this case, a faithful explanation should cause the model's confidence to increase rapidly.
\end{enumerate}

Then the area under the curve (AUC) is computed for both metrics. A lower AUC for the deletion curve and a higher AUC for the insertion curve indicate better correctness. Together, these complementary metrics confirm that the highlighted regions drive the model's output, even though they do not assess whether those regions are semantically meaningful to a human reader.

\subsection{Local Fidelity}
To evaluate the interpretable linear regression model, we rely on the coefficient of determination, $R^2$~\cite{Montgomery2021}. In this context, $R^2$ serves as a measure of local fidelity, reflecting how accurately the explanations capture the local behaviour of the underlying model~\cite{tan_glime_2023}. A $R^2$ closes to 1 indicates a high local fidelity, meaning the surrogate model accurately reflects the underlying model's behaviour. An $R^2$ close to 0 indicates a low local fidelity, meaning the explanation fails to capture the local behaviour and is therefore unreliable.

\section{Results and Discussion}\label{sec:results}
We now present the empirical results of our segmentation study. We analyze how segmentation affects LIME consistency and convergence, assess local faithfulness, and show how document-aware segmentation helps reveal shortcuts in RVL-CDIP.

\subsection{Segmentation: Main Driver of Consistency}
Across all tested settings, segmentation emerges as the dominant factor affecting LIME consistency. Document-aware segmentations, in particular \textbf{BBs only} and \textbf{Grid~$4{\times}4$ w/o BBs}, consistently outperform standard superpixel methods. As shown in Table~\ref{tab:segmentation_lime_evaluation}, Quickshift produces the least stable explanations (average Spearman consistency of $0.432$), while SLIC offers a partial improvement yet remains inferior to the document-aware alternatives, which reach substantially higher consistency scores of $0.933$ and $0.975$, respectively. These results indicate that explanation stability depends strongly on how interpretable regions align with the inherent structure of the document.


\begin{figure}[ht]
    \centering
    \includegraphics[width=.9\linewidth]{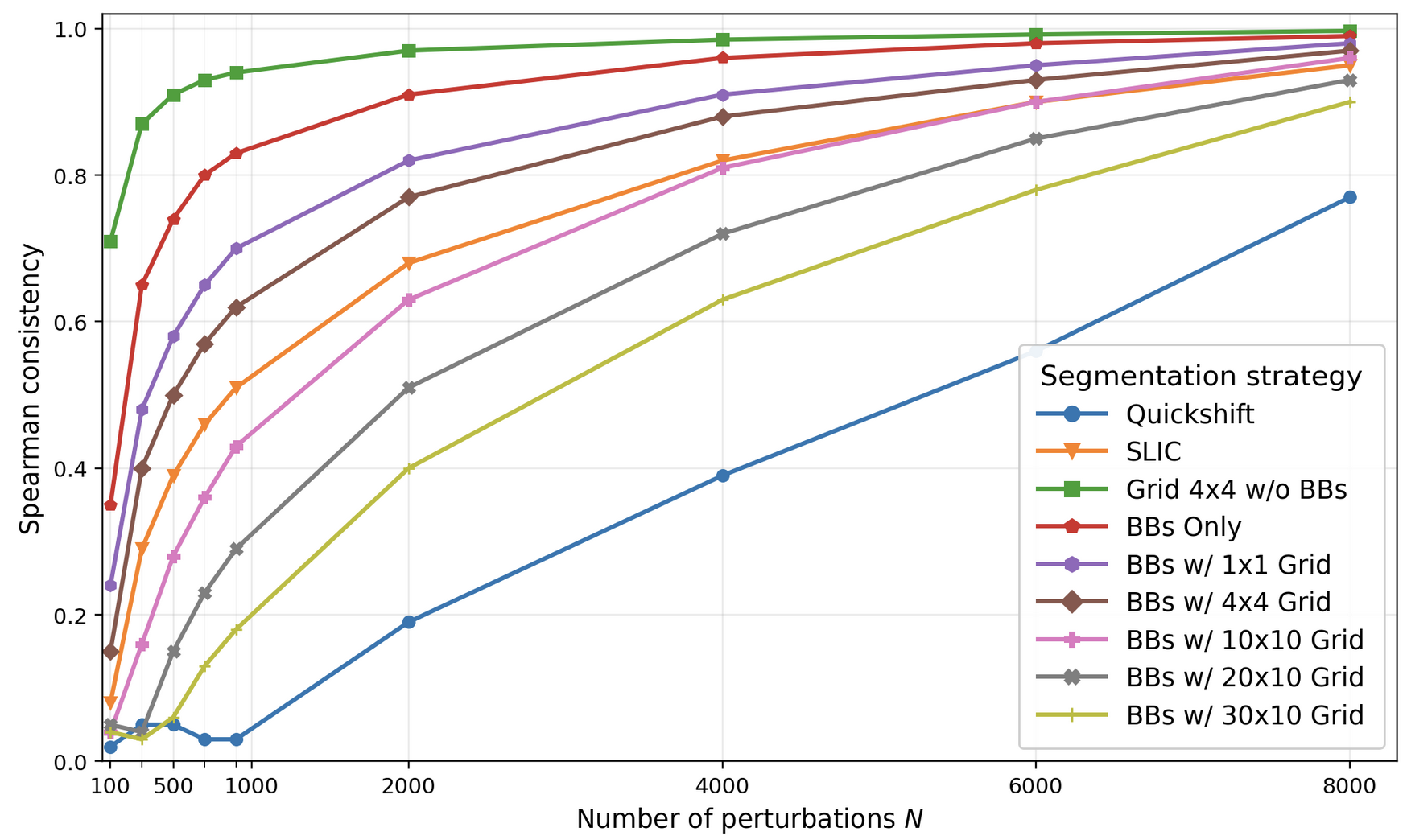}
    \caption{Spearman consistency convergence vs. number of perturbations $N$.}
    \label{fig:spearman_consistency_regime}
\end{figure}

\begin{figure}[ht]
    \centering
    \includegraphics[width=0.9\linewidth]{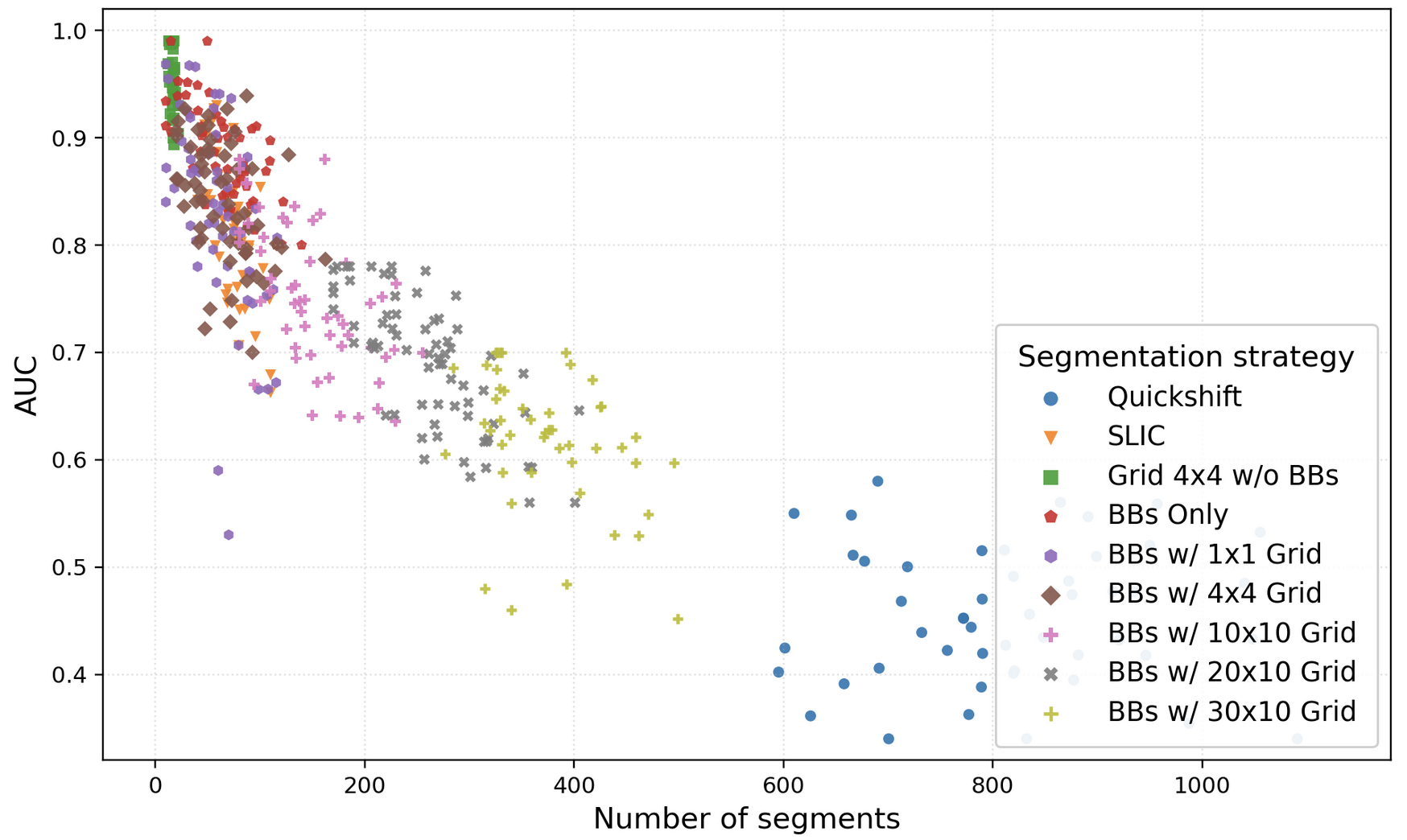}
    \caption{Normalized area under the Spearman consistency curve as a function of the number of segments in the image. Each point represents one image.}
    \label{fig:number_segments_consistency}
\end{figure}

 A clear convergence pattern is observed. Consistency improves as the number of perturbations increases, as shown in Figure~\ref{fig:spearman_consistency_regime}, but the rate of convergence is strongly linked to the segmentation strategy. Segmentations with fewer and more structured regions reach high consistency with fewer samples, whereas high-dimensional segmentations require substantially more perturbations before explanations stabilise. Figure~\ref{fig:number_segments_consistency} supports this interpretation, showing that the normalised area under the consistency curve decreases as the number of segments grows.

This behaviour is explained by the structure of the interpretable space itself. Superpixel-based methods fragment text and layout into many irregular components, increasing the dimensionality of the perturbation space and making repeated LIME runs more sensitive to random sampling. Document-aware segmentations, by contrast, produce regions that are more semantically coherent and often lower-dimensional, allowing the local surrogate model to estimate feature importance more reliably. However, increasing grid granularity progressively reduces stability: fine-grained \textbf{BBs w grid} variants such as \(10\times10\), \(20\times10\), and \(30\times10\) become less stable than SLIC. These findings support viewing LIME as a local estimation problem whose sample complexity is determined by the choice of segmentation.

\subsection{Explanation Quality: Correctness \& Fidelity}
Beyond consistency, document-aware segmentations yield substantially better explanation quality, as measured by both correctness and local fidelity metrics (Table~\ref{tab:segmentation_lime_evaluation}).

\paragraph{Correctness.} The \textbf{BBs only} strategy emerges as the superior approach, achieving the highest insertion AUC of $0.764$ and the lowest deletion AUC of $0.031$. This divergence between insertion and deletion scores confirms that the textual regions isolated by OCR are the primary drivers of the model's output. The Grid $4\times4$ w/o BBs method also captures a significant signal, yielding the second highest insertion score of $0.589$. However, its higher deletion score of $0.389$ suggests that coarse grids may inadvertently bundle predictive features with uninformative background noise.

Conversely, superpixel-based segmentations fail significantly to produce faithful explanations for document images. Quickshift records the lowest insertion AUC ($0.080$), implying that these zones are not sufficiently decisive in the decision-making process when progressively displayed to the model. As for the SLIC method, it yields an insertion score that falls in the middle of the document-aware segmentation algorithms. However, both SLIC and Quickshift yield low deletion metrics ($0.073$ and $0.089$, respectively), indicating that model confidence drops when these zones are progressively masked.

\begin{figure*}[htbp] 
    \centering
    \begin{subfigure}{0.23\textwidth}
        \includegraphics[width=\linewidth]{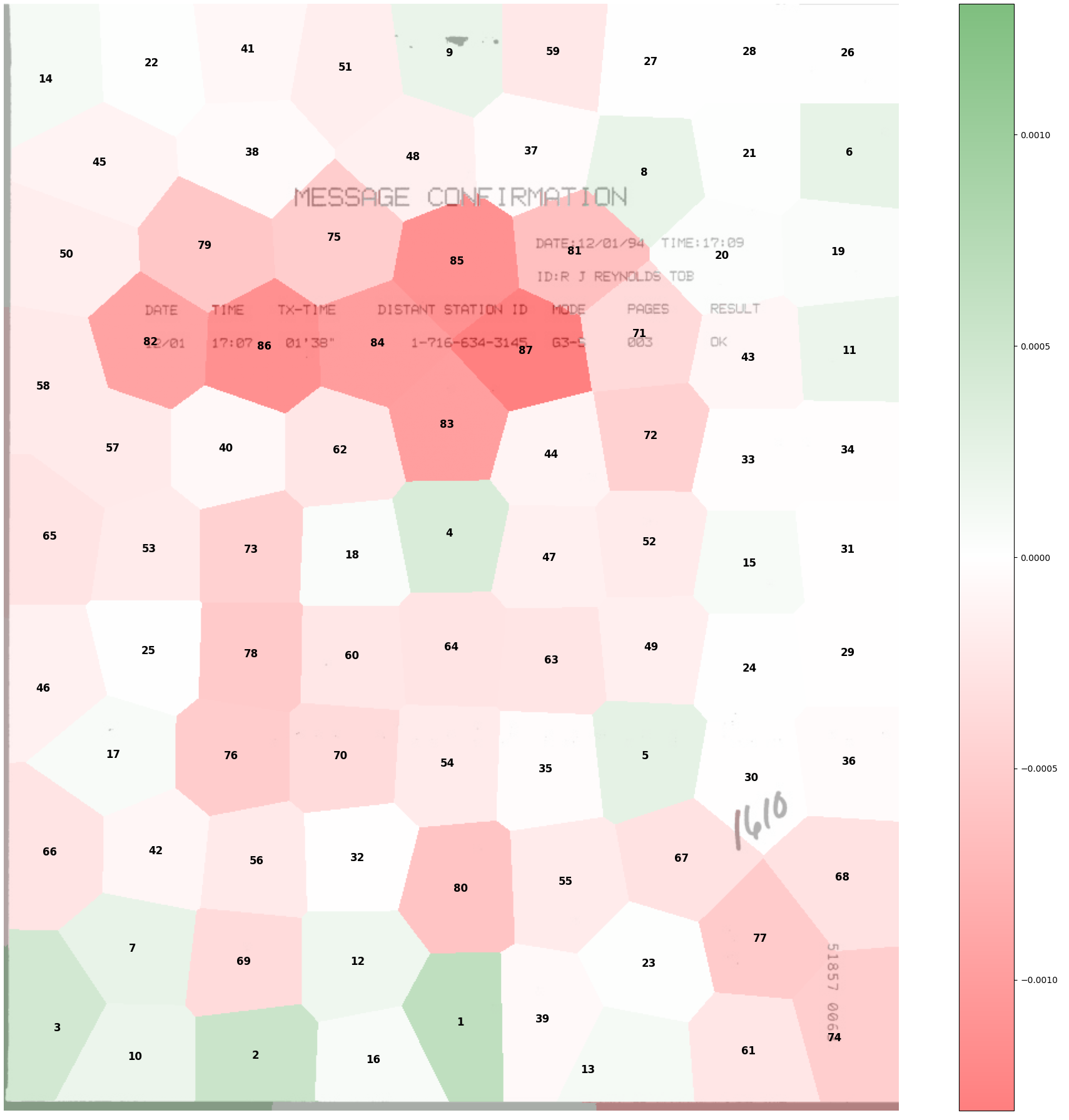}
        \caption{Doc.~1 SLIC}
        \label{fig:4slic}
    \end{subfigure}\hfill
    \begin{subfigure}{0.23\textwidth}
        \includegraphics[width=\linewidth]{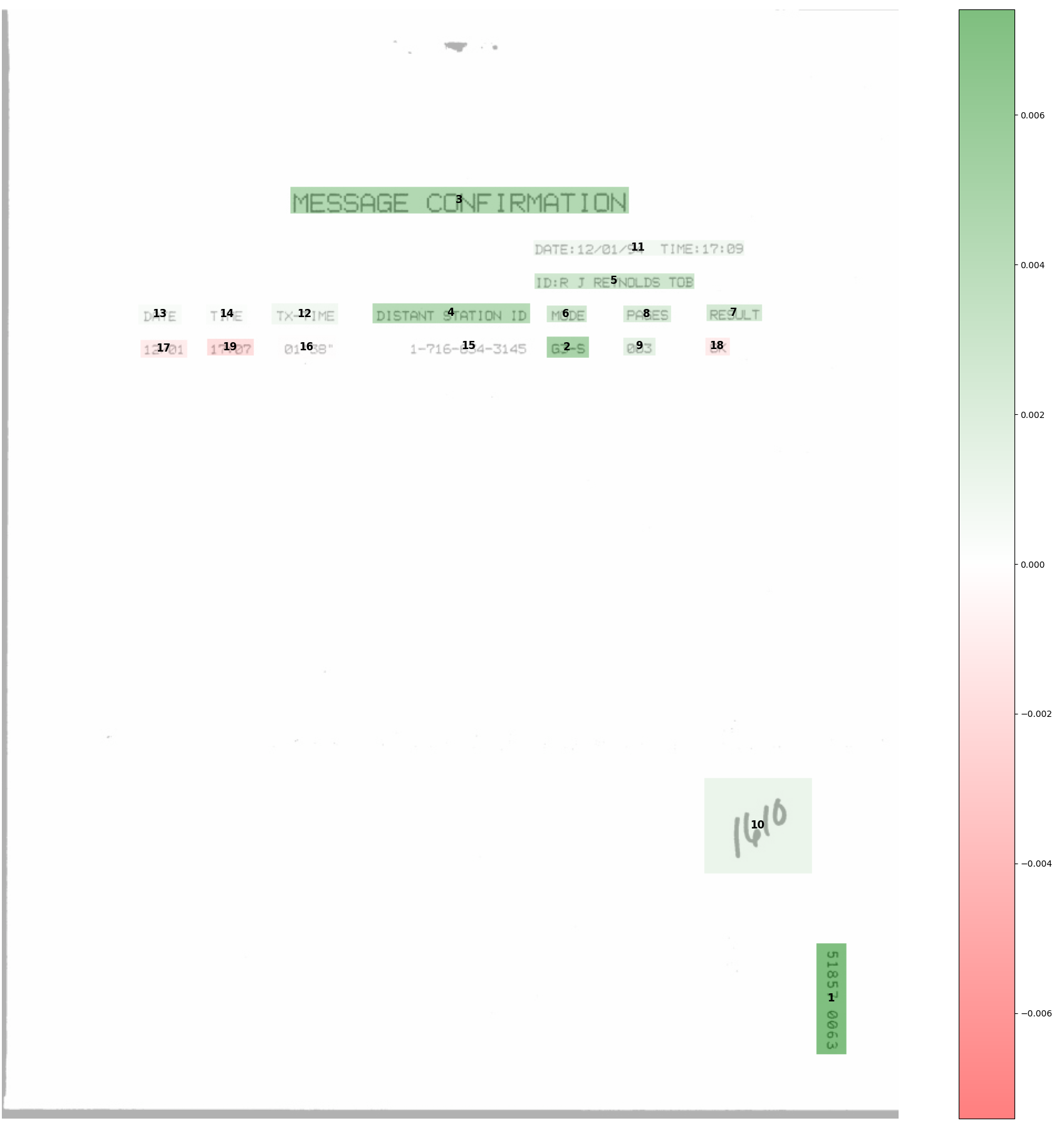}
        \caption{Doc.~1 BBs only}
        \label{fig:4bbs}
    \end{subfigure}\hfill
    \begin{subfigure}{0.23\textwidth}
        \includegraphics[width=\linewidth]{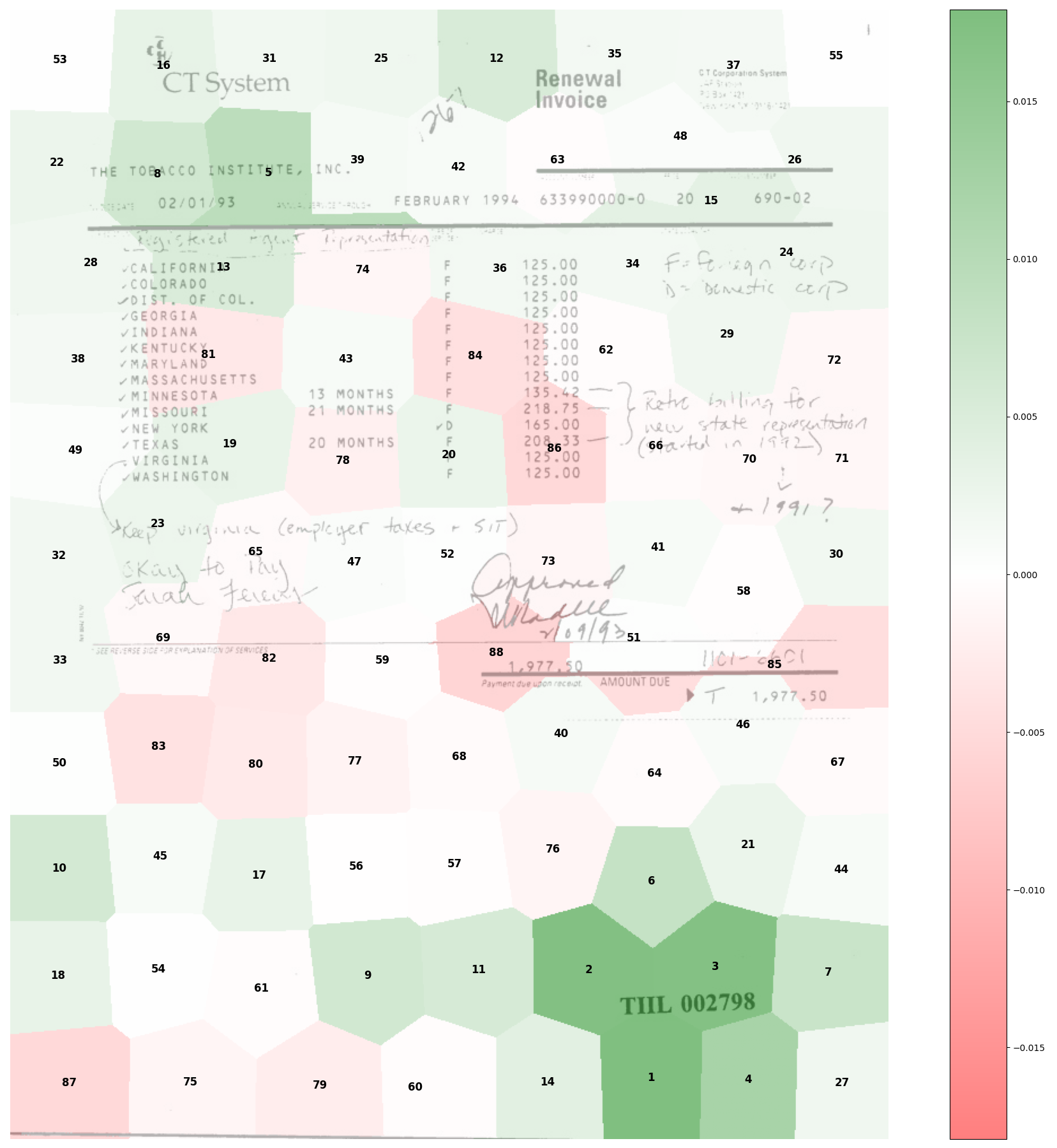}
        \caption{Doc.~2 SLIC}
        \label{fig:55slic}
    \end{subfigure}\hfill
    \begin{subfigure}{0.23\textwidth}
        \includegraphics[width=\linewidth]{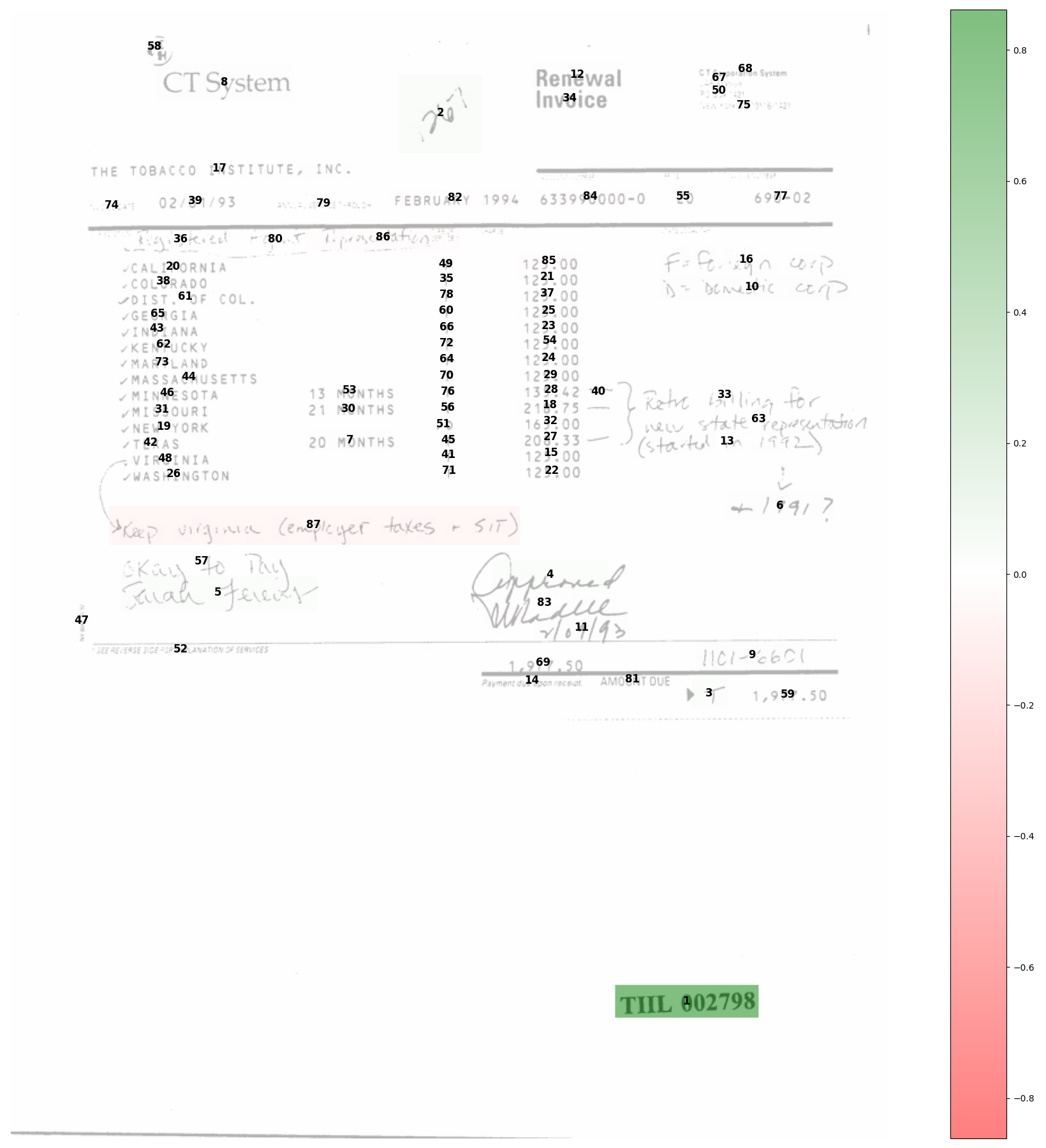}
        \caption{Doc.~2 BBs only}
        \label{fig:55bbs}
    \end{subfigure}


    \caption{Examples of bias discovery in RVL-CDIP. The figure compares SLIC and BBs-only explanations across two distinct scenarios. Document 1 (Figures~\ref{fig:4slic} and~\ref{fig:4bbs}) illustrates a case where only the BBs-only approach successfully surfaces the ID code, while SLIC dilutes its importance due to coarse superpixels. Conversely, Document 2 (Figures~\ref{fig:55slic} and~\ref{fig:55bbs}) demonstrates a case where the model's reliance on the ID code is so strong that even SLIC partially captures it, though the BBs-only method isolates it most distinctly.}
    \label{fig:biais-example-discoveries}
\end{figure*}

\paragraph{Local fidelity.} The document-aware segmentations yield substantially higher local fidelity than superpixel-based methods. The \textbf{BBs w/ $1\times1$ Grid} and \textbf{BBs only} configurations achieve the highest fidelity scores ($0.571$ and $0.557$), followed by \textbf{Grid $4\times4$ w/o BBs} ($0.508$). Superpixel-based segmentations struggle to model the local decision boundary: Quickshift yields an $R^2$ of $0.389 \pm 0.040$, while SLIC drops to $0.218 \pm 0.142$. These findings reinforce the consistency results. Because superpixel algorithms group pixels by local visual similarity rather than structural regularities, they often fragment text and layout into irregular components. Consequently, the interpretable units used by LIME do not correspond to meaningful document components, leading to structurally unrealistic perturbations. Document-aware segmentations instead use regions aligned with textual zones and spatial configurations, yielding explanations that are more faithful to the underlying model's behaviour.

\subsection{Bias Discovery}

Recent work has exposed a critical bias in RVL-CDIP. \cite{saifullah_reality_2024} showed that many documents contain class-specific identification numbers, which deep learning models often rely on instead of meaningful visual features, affecting 25\%–50\% of predictions. \cite{larson_spurious_2025} further confirmed that these ID codes function as spurious shortcuts: state-of-the-art models, including vision transformers, suffer substantial accuracy drops when the codes are obscured. Importantly, \cite{saifullah_reality_2024} also found that standard LIME explanations fail to reveal this bias, because their perturbation mechanism is too coarse to capture such fine-grained artifacts and often produces explanations that are difficult to interpret.

In contrast, we demonstrate that our method successfully identifies this bias, highlighting its ability to reveal shortcuts that conventional explainability approaches miss. Figure~\ref{fig:biais-example-discoveries} illustrates a comparison between the \textbf{SLIC} and \textbf{BBs only} algorithms across two document images, each represented in a separate column. 

For the first image (Figures~\ref{fig:4slic} and~\ref{fig:4bbs}), the two approaches assign markedly different importance to the identification number. With SLIC (Figure~\ref{fig:4slic}), the identifier does not emerge as a decisive region; the surrounding segments are highlighted in red, indicating that the model's attention is distributed across broader, less discriminative areas. By contrast, with \textbf{BBs only} (Figure~\ref{fig:4bbs}), the identifier is assigned the highest coefficient in the explanation vector, suggesting that the model relies on it predominantly for its prediction. We attribute this discrepancy to the inherently coarse granularity of SLIC superpixels: because the identifier spans only a small spatial region, it tends to be merged with surrounding background areas, thereby diluting its individual contribution to the explanation. 
For the last image (Figures~\ref{fig:55slic} and~\ref{fig:55bbs}), however, both algorithms successfully detect the identifier as a critical region. Even with SLIC, where the identifier is distributed across multiple segments, it still receives notable attribution. More importantly, the \textbf{BBs only} explanation (Figure~\ref{fig:55bbs}) assigns this region a substantially higher coefficient than all others, strongly suggesting that the model relies almost exclusively on this zone to classify the document. In this case, because the model's decision is so heavily anchored to the identifier, the background noise introduced by SLIC's coarser segmentation does not negatively affect the detection of the shortcut.

To more precisely characterize this observation, we perform a statistical analysis on 60 images from the dataset. Although a larger sample size would be preferable, we limit the analysis to 60 images because extracting, aligning, and assessing region-specific explanations for both methods demands substantial computational resources and extensive manual verification. Among these, \textbf{BBs only} assigns the identification code as the single most important region in the explanation for 15 documents. When examining the corresponding explanations produced by \textbf{SLIC} for those same 15 documents, the identification code region appears within the top 5 most important segments in only 5 cases. For the remaining 10 documents, the rank of the identification code under SLIC is substantially degraded, with an average rank of approximately 39 and values reaching as high as 71 out of all segments. This means that in the majority of cases, SLIC pushes the identification code far down in the explanation ranking, effectively hiding it from the analyst's attention. This quantitative evidence corroborates our earlier qualitative findings and further demonstrates that the coarse granularity of SLIC superpixels causes the identification code to be absorbed into surrounding background regions, artificially suppressing its importance in the final explanation and rendering the shortcut undetectable.



\section{Conclusions}
In this work, we showed that segmentation is one of the main factor in the quality of LIME explanations for document image classification. Document-aware segmentations, especially \textbf{BBs only}, produce more consistent, correct, and locally faithful explanations than standard superpixel methods. They also converge with fewer perturbations by defining a more coherent and lower-dimensional interpretable space. Finally, we showed that document-aware segmentation helps reveal RVL-CDIP shortcut behaviour based on document identification codes, which LIME with superpixel segmentation often obscures. These findings suggest that post-hoc explanations should use interpretable representations aligned with the structure of the input domain. For future work, we plan to extend this analysis to additional document classifiers, including layout-aware and other architectures, and go beyond the RVL-CDIP benchmark. 

\section*{Acknowledgments}
This research was funded by Itesoft, the ANRT CIFRE PhD program, and ACTUADA (No. 2022-2021-17014610) project funded by the Nouvelle-Aquitaine Region, France. It benefited from the computing resources of the L3i laboratory at the University of La Rochelle, funded by the French government and the Nouvelle-Aquitaine Region.


\bibliographystyle{named}
\bibliography{ijcai26}

\end{document}